\documentclass[acmsmall, nonacm]{acmart}

\AtBeginDocument{%
  \providecommand\BibTeX{{%
    \normalfont B\kern-0.5em{\scshape i\kern-0.25em b}\kern-0.8em\TeX}}}

\usepackage{hyperref}
\begin{document}

\title{Factorization of Fact-Checks for Low Resource Indian Languages}

\author{Shivangi Singhal, Rajiv Ratn Shah, Ponnurangam Kumaraguru}
\email{(shivangis, rajivratn, pk)@iiitd.ac.in}
\affiliation{%
  \institution{IIIT-Delhi}
}








\renewcommand{\shortauthors}{Singhal, et al.}

\begin{abstract}
The advancement in technology and accessibility of internet to each individual is revolutionizing the real time information. The liberty to express your thoughts without passing through any credibility check is leading to dissemination of fake content in the ecosystem. It can have disastrous effects on both individuals and society as a whole. The amplification of fake news is becoming rampant in India too. Debunked information often gets republished with a replacement description, claiming it to depict some different incidence. To curb such fabricated stories, it is necessary to investigate such deduplicates and false claims made in public. The majority of studies on automatic fact-checking and fake news detection is restricted to English only. But for a country like India where only \emph{10\%} of the literate population speak English, role of regional languages in spreading falsity cannot be undermined. In this paper, we introduce \textbf{FactDRIL}: the first large scale multilingual \textbf{Fact}-checking \textbf{D}ataset for \textbf{R}egional \textbf{I}ndian \textbf{L}anguages. We collect an exhaustive dataset across \emph{7} months covering \emph{11} low-resource languages. Our propose dataset consists of \emph{9,058} samples belonging to English, \emph{5,155} samples to Hindi and remaining \emph{8,222} samples are distributed across various regional languages, \emph{i.e.} Bangla, Marathi, Malayalam, Telugu, Tamil, Oriya, Assamese, Punjabi, Urdu, Sinhala and Burmese. We also present the detailed characterization of three M’s (multi-lingual, multi-media, multi-domain) in the FactDRIL accompanied with the complete list of other varied attributes making it a unique dataset to study. Lastly, we present some potential use cases of the dataset.  We expect this dataset will be a valuable resource and serve as a starting point to fight proliferation of fake news in low resource languages.


\end{abstract}



\maketitle

\begin{figure}
  \centering
  \includegraphics[width=\linewidth]{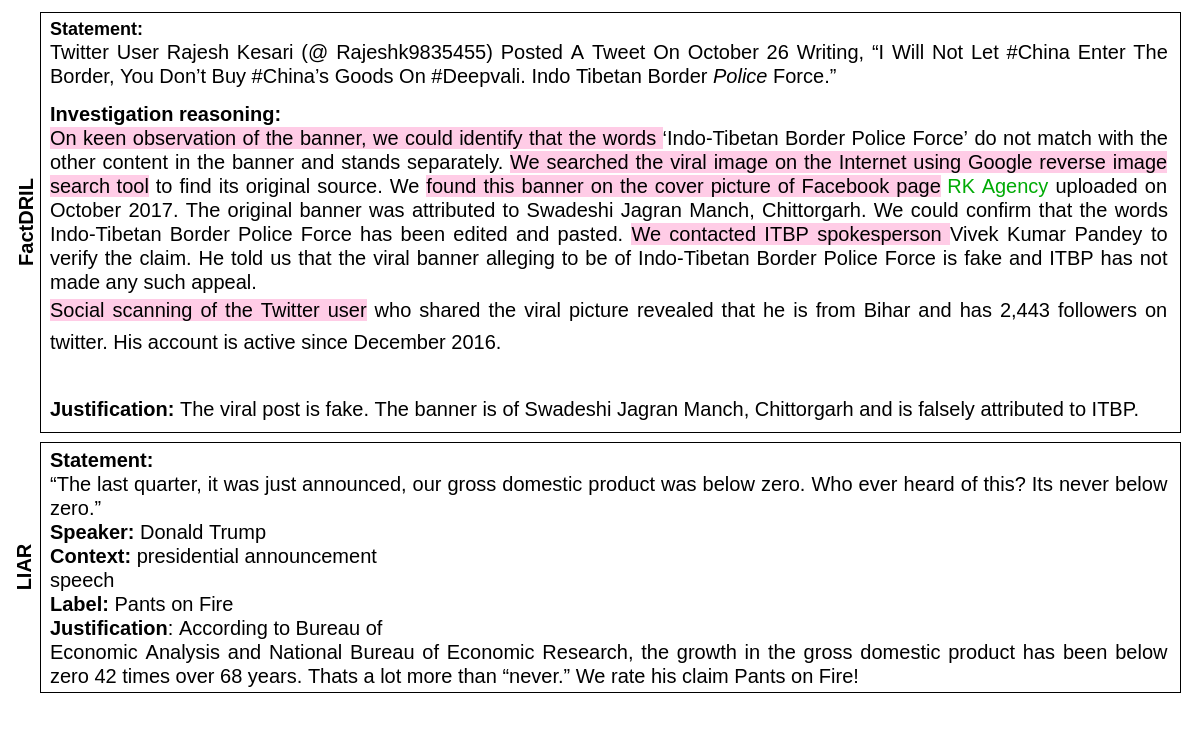}
  \caption{An excerpt from our proposed FactDRIL and LIAR \cite{liar} is shown. The pink highlighted text shows the investigation steps adopted by fact-checkers. Though the justification (veracity\_reasoning) in both the dataset is present but the \emph{investigation\_reasoning} in FactDRIL gives minute details of the fact-checking process. This attribute is exclusive of FactDRIL and is not present in any of the existing fact-checking datasets.}
  \label{invest_reasoning}
\end{figure}

\section{Introduction}
Fake news is spreading like wildfire. All the major happenings around the globe like 2016 U.S. elections \cite{US}, global pandemic (COVID-19) \cite{covid, pal} and 2019 Indian General elections \cite{indian_elections} have been heavily affected by the proliferation of infodemic on social and mainstream media. To limit the escalation of fake news, fact-checkers are constantly making an effort to find out the authenticity behind the unverified claims. Fact-checking is generally defined as a process to find out the veracity behind such viral claims. The conclusion drawn is always backed up by evidence and logical reasonings \cite{ACL_2014}. 
Fake news is a global concern but majority of the automatic fact-checking \cite{ACL_2014,emergent, liar, liarplus, claimkg, multifc} and fake news detection \cite{kai,eann, mvae,safe, spotfake, spotfake_plus} solutions have been designed for English, a language predominantly used in web-based social networking. Though that is good on a global scale, but when it comes to India, it fails due to following reasons, \emph{(i)} people speak diversified range of languages, \emph{(ii)} due to housing a number of languages, the communication in English is bit problematic because out of the 74\% literates in the country, only 10\% can read English\footnote{https://en.wikipedia.org/wiki/2011\_Census\_of\_India}. Recently, 
Rasmus Kleis Nielsen, director at Reuters Institute for the Study of Journalism, also addresses this issue in his interview\footnote{https://www.thequint.com/news/india/media-coverage-disinformation-in-india-interview-rasmus-nielsen} stating that, the problems of disinformation in a society like India might be more sophisticated and tougher than they are within the West. We wish to understand the menace of misinformation in Indian territory where false information, manipulative photos and deduplicate news reappear on the online ecosystem time and again. To the best of our knowledge, there is no existing fact-checking data for Indian languages. This has caused hindrance in devising solutions for automatic fact-checking, which can negatively affect the large group of population.

In this paper, we aim to bridge the gap by proposing \textbf{FactDRIL}: the first large scale multilingual \textbf{Fact}-checking \textbf{D}ataset for \textbf{R}egional \textbf{I}ndian \textbf{L}anguages. Our contribution can be summarized as follows:
\begin{itemize}
    \item  We curated \emph{22,435} samples from the eleven Indian fact-checking website certified with IFCN ratings. The samples are in the various low-resource languages: Bangla, Marathi, Malayalam, Telugu, Tamil, Oriya, Assamese, Punjabi, Urdu, Sinhala and, Burmese. We release the dataset that can act as a prime resource for building automatic fact-checking solutions for Indian ecosystem.\footnote{https://bit.ly/3btmcgN} 
    
    \item We introduced an attribute in the feature list termed as \emph{investigation\_reasoning}. This attribute provides an explanation of the intermediate steps performed by fact-checkers to conclude the veracity of the unverified claim. This is important to study because it will help us to dig into the fact-checking mechanism and propose solutions to automate the process. We discuss in detail the use-case of this curated feature and  the methodology designed to excavate it from the crawled unstructured data dump.
    
\end{itemize}

\section{Related Work}
There have been several datasets released in past that focused on fact-checking. An overview of the core fact-checking datasets is given in Table \ref{tab: lit_review_fc}.

The very first effort towards this direction was made by \cite{ACL_2014} in 2014. The paper released a publicly available dataset that consist of  sentences fact-checked by journalist available online. The statements were picked from  the fact-checking blog of Channel 4 and the Truth-O-Meter from PolitiFact. The statements mainly captured issues prevalent in U.S. and U.K. public life. Apart from statements, the meta-data features like, \emph{(i)} publish date, \emph{(ii)} speaker, \emph{(iii)} fine-grained label associated with the verdict and, \emph{(iv)} URL were also collected.

Another study was done in 2016 by \cite{emergent}. In this paper, the data was  collected from numerous sources including rumor sites and Twitter handles. The news pertaining to the world, U.S. national and technology were captured. For each claim, journalist would search for the articles that are either in \emph{support}, \emph{against} or \emph{observing} towards the claim. The final dataset consist of claims with corresponding summarized headings by the journalist and associated veracity label with the final verdict on the claimed statement.

Both the previously mentioned datasets were quite small in numbers. To overcome this drawback, the LIAR dataset was introduced by \cite{liar} in 2017. It consists of around 12.8K short statements curated from Politifact website. It mainly contain samples collected from a variety of sources including, TV interviews, speeches, tweets and debates. The samples cover wide range of issues ranging from the economy, health care, taxes to
elections. The samples were annotated for truthfulness, subject, context, speaker, state, party, and prior history. For truthfulness, the dataset was equally distributed into six labels: pants-fire, false, mostly false, half-true, mostly-true, and true. In 2018, Alhindi \emph{et al.} \cite{liarplus} proposed LIAR-PLUS, an extended version of the LIAR dataset. For each sample, human justification for the claim was automatically extracted from the fact-checking article. It was believed that justification combined with extracted features and meta-data would boost the performance of classification models. Another dataset that came into existence in 2018 was FEVER \cite{fever}. It consists of 185,445 claims that were not naturally occurring but were generated by altering sentences extracted from Wikipedia.

Later in 2019, two new functionalities \emph{i.e.} evidence pages and knowledge graph triples were introduced to fact-checking data that resulted in overall improvement of the accuracy. Augenstein \emph{et al.} \cite{multifc} presented the largest dataset on fact-checking that had \emph{34,918} claims collected from \emph{26} fact-checking websites in English listed by Duke Reporters Lab and on fact-checking Wikipedia page. The prime benefit introduced was the \emph{10} relevant evidence pages per claim. Other features included claim,
label, URL, reason for label, categories, speaker, fact-checker, tags, article title, publication, date, claim date, and the full text associated with claim. Authors in \cite{claimkg} argue the need of labelled ground truth and other features to perform supervised machine learning methods. and emphasis on the requirement of a huge storage space. To solve the problem, they proposed ClaimsKG, a knowledge graph of fact-checked claims. It enables structured queries for features associate with the claim. It is a semi-automated method that gather data from fact-checking websites and annotate claims and their corresponding entities from DBpedia.

Next, we discuss how our proposed dataset- \emph{FactDRIL} is different from the existing fact-checking datasets.

\begin{table}[]
\caption{Comaprison of dfferent fact-checking dataset. The table clearly shows that the proposed dataset is rich in terms of linguistic diversity and features. It also captures the information in the form of multiple modalities and from differnent domains. Values in ( ) denote the number of meta-features extracted in the each dataset. \emph{`NA'}  means that the information is not explicitly mentioned.}
\label{tab: lit_review_fc}
\begin{tabular}{|l|l|l|l|l|l|l|l|}
\hline
                                                                    & \textbf{Region}                                        & \textbf{\begin{tabular}[c]{@{}l@{}}Moda-\\ lity\end{tabular}}                               & \textbf{Domain}                                                                                             & \textbf{\begin{tabular}[c]{@{}l@{}}Multi-\\ Ling-\\ ual\end{tabular}} & \textbf{\begin{tabular}[c]{@{}l@{}}Investi-\\ gation \\ steps\end{tabular}} & \textbf{\# Claims} & \textbf{\begin{tabular}[c]{@{}l@{}}Claim \\ Sources\end{tabular}}                                                                    \\ \hline
ACL 2014 \cite{ACL_2014}                                                            & \begin{tabular}[c]{@{}l@{}}U.S.A, \\ U.K.\end{tabular} & text                                                                                        & public life                                                                                                 & No                                                                    & No                                                                          & 106 (4)            & \begin{tabular}[c]{@{}l@{}}Politifact, \\ Channel\\  4 News\end{tabular}                                                             \\ \hline
Emergent \cite{emergent}                                                            & U.S.A                                                  & text                                                                                        & \begin{tabular}[c]{@{}l@{}}world, \\ U.S.-\\ national, \\ technology\end{tabular}                           & No                                                                    & No                                                                          & 300 (4)            & \begin{tabular}[c]{@{}l@{}}snopes.com, \\ Twitter \\ handle:\\ @Hoaxalizer\end{tabular}                                              \\ \hline
LIAR \cite{liar}                                                                & U.S.A                                                  & text                                                                                        & \begin{tabular}[c]{@{}l@{}}economy, \\ health care, \\ taxes, \\ elections\end{tabular}                     & No                                                                    & No                                                                          & 12800 (5)          & \begin{tabular}[c]{@{}l@{}}Politifact, \\ Channel\\  4 News\end{tabular}                                                             \\ \hline
LIAR-plus \cite{liarplus}                                                           & U.S.A                                                  & text                                                                                        & \begin{tabular}[c]{@{}l@{}}economy, \\ health care, \\ taxes, \\ elections\end{tabular}                     & No                                                                    & No                                                                          & 12836 (5)          & \begin{tabular}[c]{@{}l@{}}Politifact, \\ Channel \\ 4 News\end{tabular}                                                             \\ \hline
Multi-FC \cite{multifc}                                                            & NA                                                     & text                                                                                        & NA                                                                                                          & No                                                                    & No                                                                          & 36534 (12)         & \begin{tabular}[c]{@{}l@{}}Duke \\ Reporters \\ Lab,\\ Fact \\ Checking \\ Wikipedia \\ page\end{tabular}                            \\ \hline
ClaimsKG \cite{claimkg}                                                            & NA                                                     & text                                                                                        & NA                                                                                                          & No                                                                    & No                                                                          & 28383 (9)          & \begin{tabular}[c]{@{}l@{}}AfricaCheck \\ FactScan,\\ PoilitiFact,\\ Snopes, \\ Check your\\ fact,\\ Truth or\\ fiction\end{tabular} \\ \hline
\textbf{\begin{tabular}[c]{@{}l@{}}FactDRIL\\(Proposed\\Dataset)\end{tabular}} & India                                                  & \begin{tabular}[c]{@{}l@{}}text, \\ image,\\ video,\\  social \\ media \\ post\end{tabular} & \begin{tabular}[c]{@{}l@{}}health, \\ society,\\ religion, \\ politics, \\ world, \\ elections\end{tabular} & Yes                                                                   & Yes                                                                         & 22435 (22)         & \begin{tabular}[c]{@{}l@{}}All IFCN \\ rated Indian \\ fact-checking \\ websites\end{tabular}                                        \\ \hline
\end{tabular}
\end{table}


\section{What is new in the proposed dataset?}
The dataset presented in the paper is unique due to following reasons:
\begin{itemize}

\item Misinformation in India: Till date, various datasets proposed in the fake news research, consist of news samples pertaining to  USA and its neighbouring countries \cite{breaking,golbeck,torabi, kai,eann, mvae,safe,spotfake,spotfake_plus}. We believe that studying disinformation in India will be more challenging than West due to, \emph{(i)} low literacy rates that not only make it hard to induce decision-making ability in an individual, but add to an explosion of fake news and divisive propaganda, \emph{(ii)} Most Indians tend to trust messages from family and friends\footnote{https://qz.com/india/1459818/bbc-study-shows-how-indians-spread-fake-news-on-whatsapp/}. This means that content gets forwarded without any checks, further driving misinformation in the social sphere, \emph{(iii)} majority of population access news primarily through WhatsApp\footnote{https://www.latimes.com/world/la-fg-india-whatsapp-2019-story.html} where information rarely gets checked for validity and authenticity.

    \item Multilingual information: 
   The 2011 Census of India\footnote{https://en.wikipedia.org/wiki/Multilingualism\_in\_India} shows that the languages by highest number of speakers (in the decreasing order) are as follows: Hindi, Bengali, Marathi, Telugu, Tamil, Gujarati, Urdu, Odia, Malayalam and Punjabi. Whereas, only 10.67\% of the total population of India converse in English. Though current datasets are in English but the above statistics explicitly indicates a need to shift the study of fake news from English to other languages too.
    
    \item Investigation reasoning: With this dataset, we present a detailed explanation of how the investigation was carried out by manual labourers in concluding the truthfulness of the viral news. The proposed attribute is totally different from the \emph{veracity reasoning} label present in the fact-checking datasets \cite{liar, liarplus}. The primary difference is that the former attribute  explains the intermediate steps performed during manual intervention whereas the latter attribute concludes the veracity of unverified claim with a reasoning without emphasising on how they searched for that reasoning. Figure \ref{invest_reasoning} explains the difference between the two. We believe such an information will be helpful in preparing solutions to automate the manual fact-checking efforts.

    \item Multimedia and multi-platform information: Fake news can be published in any form and on any social and mainstream platform. The curated dataset incorporates the information about  media (images, text, video, audio or social media post) used in fake news generation  and the medium (Twitter, Facebook, WhatsApp and Youtube) used to populate it in the surroundings.
    
    \item Multi-domain information: Previous fact-checking dataset as shown in Table \ref{tab: lit_review_fc} covers information in certain domain only. For example, Emergent \cite{emergent} only captures the national, technological and world related happening in U.S.A whereas \cite{liar, liarplus} includes health, economy and election related issues. This is due to the fact that different fact-checking websites focuses on capturing news of specific genre. Since in our proposed dataset, we crawled information from all the fact-checking websites that exist in the country. This gives us a leverage to encapsulate all the happenings around the territory making it a rich dataset to study.
    

\end{itemize}

To sum up, in our perception, this is the first dataset introduced in the fake news domain that  collects information from Indian territory. The detailed characterization of three M's (multi-lingual, multi-media, multi-domain) in the dataset accompanied with veracity reasoning and other varied attributes makes it a unique dataset to study.


\section{Data Collection}
\begin{figure}[h]
  \centering
  \includegraphics[width=\linewidth]{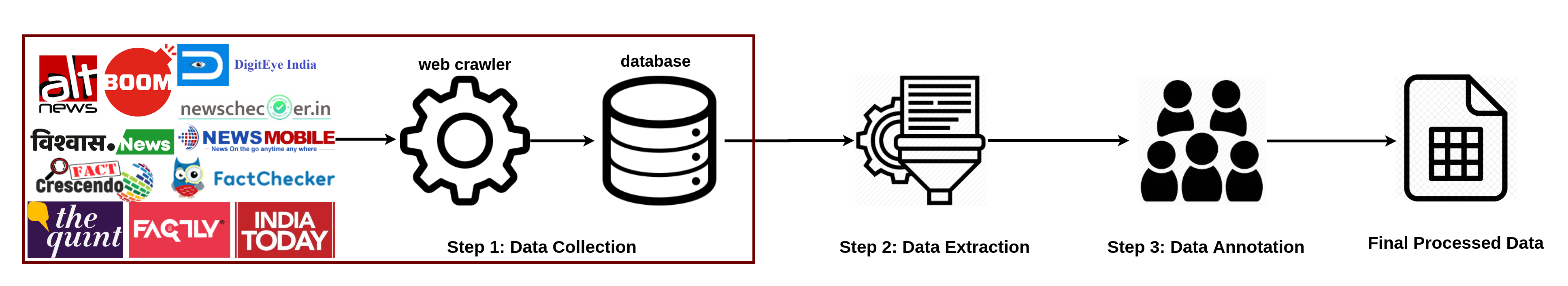}
  \caption{Our proposed dataset curation pipeline. \emph{Step 1} describes the data collection process followed by \emph{Step 2} describing the Data Extraction and \emph{Step 3} of Data Annotations.}
  \label{data_curation}
\end{figure}

In our opinion, we have curated, the first large scale multilingual Indian fact-checking data. Figure \ref{data_curation} shows the complete data curation process. In this section, we focus on the first step of the curation process \emph{i.e.} the data collection.

Though fact-checking services play a pivotal role in combating misinformation, little is known about whether users can rely on them or not. To corroborate trust among audience, fact-checking services should endeavour transparency in their processes, as well as in their organizations, and funding sources. With an objective to look out for trusted Indian fact-checking websites, we came across International Fact Checking Network (IFCN).
Next we discuss in detail about IFCN, its measuring criteria and sources chosen for preparing the final dataset.

\subsection{International Fact-Checking Network}
The International Fact-Checking Network is owned by Poynter Institute of Medical Studies located in St. Petersburg, Florida. It was set in motion on September 2015. The prime objective to establish IFCN was to bring together the fact-checkers present across the globe, under one roof. It also intent to provide a set of guidelines through the fact-checkers code of principles that are mandatory for the fact-checking organizations to follow. The code of principles is designed for agencies that actively work towards broadcasting the the correct investigation against the false claim made either on mainstream or social media platforms.

The organizations that are legally registered with an objective to routinely scrutinize the statements made by public figures and prominent institutions are generally granted the IFCN signatory. The statements can be in the form of text, visual, audio and  other formats particularly related to public interest issues. On the other hand, organizations whose opinion look influenced by the state or any other influential identity or a party are generally not admitted the grant. 

To be eligible to get an IFCN signatory, the organization is critiqued by independent assessors on \emph{31} criteria. The assessment is then finally reviewed by the IFCN advisory board to ensure fairness and consistency across the network. There are about \emph{82} Verified signatories of the IFCN code of principles among which \emph{11} are based on India. To ensure the authenticity and verifiability of the curated data, we have considered those Indian fact-checking sources that are IFCN rated verified. 

Next, we discuss the \emph{11} Indian fact-checking websites considered for our data collection process.

\subsection{Indian Verified Fact-Checking Websites}
The prime benefit of gathering data from fact-checking websites is that we can read the reasoning behind the veracity of a news sample. The detailed description of the investigation gives an useful insight to the reader about how and why the viral claim was false. With this objective in mind, we decided to collect data from the \emph{11} fact-checking websites that are on a mission to debunk fake information from the Indian ecosystem.
An overview of the fact-checking websites considered for data curation is provided in Table \ref{source_overview}. The table highlights the key features of a particular website in the form of, \emph{(i)} who formed the website, \emph{(ii)} organization establishment year, \emph{(iii)} languages debunked by the website and, \emph{(iv)} domain covered.
\begin{table}[]
\caption{An overview of fact-checking sources considered during the data collection. \emph{`NA'} means that the information is not available.}
\label{source_overview}
\begin{tabular}{|l|l|l|l|l|}
\hline
\textbf{Website}  & \textbf{Formed by}                                                                                    & \textbf{\begin{tabular}[c]{@{}l@{}}Establish \\ year\end{tabular}} & \textbf{\begin{tabular}[c]{@{}l@{}}Languages \\ supported\end{tabular}}                                                                                                  & \textbf{Domain}                                                                                             \\ \hline
Alt News          & Pratik Sinha                                                                                          & Feb 2017                                                           & English, Hindi                                                                                                                                                           & \begin{tabular}[c]{@{}l@{}}Politics, \\ Science, \\ Religion, \\ Society\end{tabular}                       \\ \hline
Boom Live         & \begin{tabular}[c]{@{}l@{}}Part of Outcue Media \\ Pvt Ltd\end{tabular}                               & Nov 2016                                                           & \begin{tabular}[c]{@{}l@{}}English, Hindi, \\ Bangla, Burmese\end{tabular}                                                                                               & General                                                                                                     \\ \hline
DigitEye India    & \begin{tabular}[c]{@{}l@{}}Part of SoftmediaHub \\ LLP\end{tabular}                                   & Nov 2018                                                           & English                                                                                                                                                                  & General                                                                                                     \\ \hline
FactChecker       & \begin{tabular}[c]{@{}l@{}}Initiative of The \\ Spending\\ Policy Research \\ Foundation\end{tabular} & Feb 2014                                                           & English                                                                                                                                                                  & \begin{tabular}[c]{@{}l@{}}General, \\ Modi-Fied\end{tabular}                                               \\ \hline
Fact Crescendo    & \begin{tabular}[c]{@{}l@{}}Part of Crescendo \\ Transcription \\ Private Limited (CTPL)\end{tabular}  & July 2018                                                          & \begin{tabular}[c]{@{}l@{}}English, Hindi, \\ Tamil, Telugu, \\ Kannada,\\ Malayalam, \\ Oriya, Assamese, \\ Punjabi, Bengali, \\ Marathi, Gujarati,\\ Urdu\end{tabular} & \begin{tabular}[c]{@{}l@{}}General, \\ Coronavirus\end{tabular}                                             \\ \hline
Factly            & Rakesh Dubbudu                                                                                        & Dec 2014                                                           & English, Telugu                                                                                                                                                          & \begin{tabular}[c]{@{}l@{}}General, \\ Coronavirus\end{tabular}                                             \\ \hline
India Today       & \begin{tabular}[c]{@{}l@{}}Part of TV Today \\ Network Limited\end{tabular}                           &                                                                    & English                                                                                                                                                                  & General                                                                                                     \\ \hline
News Mobile       & \begin{tabular}[c]{@{}l@{}}Part of World Mobile \\ NewsNetwork Pvt Ltd\end{tabular}                   & 2014                                                               & English                                                                                                                                                                  & General                                                                                                     \\ \hline
NewsChecker       & \begin{tabular}[c]{@{}l@{}}Initiative of NC Media \\ Networks Pvt Ltd\end{tabular}                    &                                                                    & \begin{tabular}[c]{@{}l@{}}English, Hindi, \\ Marathi, Punjabi, \\ Gujrati, Tamil, \\ Urdu, Bengal\end{tabular}                                                          & General                                                                                                     \\ \hline
Vishvas News      &                                                                                                       &                                                                    & \begin{tabular}[c]{@{}l@{}}English, Hindi, \\ Punjabi, Odia, \\ Assamese, \\ Gujrati, Urdu, \\ Tamil, Telugu, \\ Malayalam, \\ Marathi\end{tabular}                      & \begin{tabular}[c]{@{}l@{}}Coronavirus, \\ Politics, \\ Society, \\ World, \\ Viral,\\  Health\end{tabular} \\ \hline
The Quint-Webqoof & \begin{tabular}[c]{@{}l@{}}Owened by Gaurav \\ Mercantiles Limited\end{tabular}                       &                                                                    & English, Hindi                                                                                                                                                           & \begin{tabular}[c]{@{}l@{}}General, \\ Health\end{tabular}                                                  \\ \hline
\end{tabular}
\end{table}

\section{Data Extraction}
In this section, we discuss the schema of our proposed dataset. This is the second step of the data curation pipeline as shown in Figure \ref{data_curation}.

\textbf{Data Collection}
We list down all the authentic fact-checking sources that unfolds claims written not only in multilingual languages but lingua franca too. We set-up a data extraction system that make use of a Python library, Beautiful Soup\footnote{https://pypi.org/project/beautifulsoup4/} to extract data from web pages. Our system checks the sources for new data once in \emph{24} hours. 

In this paper, we present a study on samples curated from a time period of \emph{Decemner 2019}- \emph{June 2020}. By the end of the data curation process, we had \textbf{\emph{22,435} news samples from \emph{11} fact-checking websites}. Among them, \emph{9,058} samples belong to English, \emph{5,155} samples to Hindi and remaining \emph{8,222} samples were distributed in various regional languages \emph{i.e.} Bangla, Marathi, Malayalam, Telugu,Tamil, Oriya, Assamese, Punjabi, Urdu, Sinhala and Burmese.


\subsection{Dataset Attributes} \label{attributes}

\begin{figure}[h]
  \centering
  \includegraphics[width=\linewidth]{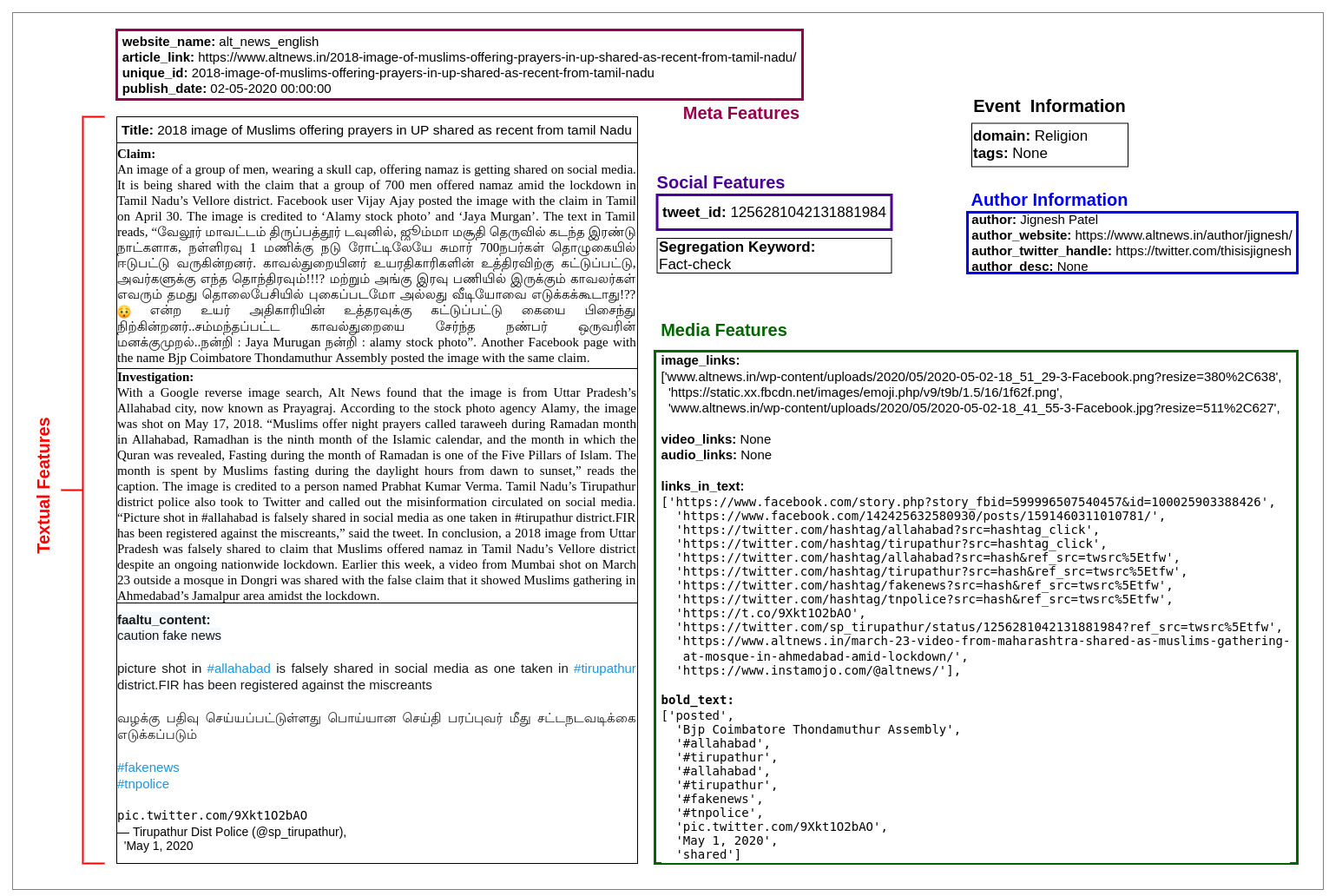}
  \caption{A excerpt from the dataset displaying different attributes present in the proposed dataset. The feature list is paced under four headers namely, \emph{meta features}, \emph{text features}, \emph{social features}, \emph{media features}, \emph{author information} and \emph{event information}. These attributes are discussed in detail in Section \ref{attributes}.}
  \label{dataset_attributes}
 
\end{figure}

We curated numerous features from the unstructured data. We have then categorized the extracted feature set into various classes like meta features, textual features, author features, media features, social features and event features. A sample showcasing all the attributes is present in Figure \ref{dataset_attributes}.

\textbf{Meta Features}
We consider those attributes as meta\_features that tells us about the sample, like \emph{website\_ name, article\_link, unique\_id, publish\_date}.

\begin{itemize}
    \item \emph{website\_name}: Denotes the name of the source from where the following sample is collected. It also gives an additional information about the language in which the fact-checked article is originally written.
    
    \item \emph{article\_link}: The attribute gives you the original link of the curated sample.
    
    \item \emph{unique\_id}: This attribute acts as the primary key for data storage.
    
    \item \emph{publish\_date}: The attribute signifies the date on which the article was published by the fact-checking websites.
\end{itemize}

\textbf{Textual Features}
A fact-checked article is generally segregated into three divisions, \emph{title of the article}, \emph{claim} and \emph{investigation}. All these together form the textual features in our proposed dataset. 
The crawled data from the website is highly unstructured. The information in the form of claim and investigation is generally present in the content part of the data. This information is extracted from the \emph{content} attribute using human intervention. 
This is  discussed in detail in Section \ref{annotation}.

\begin{itemize}
    \item \emph{title}: The title of the article.
    
    \item \emph{content}: This attribute act as the body of the article that consist of information in the form of claim and investigation.
    
    \item \emph{faaltu\_content}: Many a times, the crawled information from the website also contains the text present in the social media post (say, text present in the tweets or Facebook posts) attached as a reference within the article. Though, such text pieces will  be a hindrance  while reading the content but they might give useful information if the corresponding social media post is deleted from web. To take into account all such possibilities, we coined an attribute, \emph{faaltu\_content}\footnote{The word `faaltu' is a Hindi word that means `useless' in English.} in our proposed \emph{FactDRIL} that takes into account the text parts of the social media post attached with the article. The example of the same is shown in Figure \ref{dataset_attributes}.
    
    \item \emph{claim}: 
    This attribute gives reader a background information about \emph{what was said in the corresponding post}.
    
    \item \emph{investigation}: This attribute help readers in understanding \emph{why the fact-checkers concluded a particular post to be fake}. The whole inspection process is discussed in detail with tools and technology used for its exploration.
    
    \item  Other features like \emph{ claim\_title, investigation\_title, post\_summary} are present in some samples. These attributes give one line summary of the \emph{claim, investigation and complete article}. Such attributes are useful when the reader intends to take a quick look at the article to find out questions like, \emph{What was claimed, who was claimant and how it was investigated to be false}.

\end{itemize}

\textbf{Author Information}
This set of attributes showcase information about the people who are involved in fact-checking.

\begin{itemize}
    \item \emph{author\_name}: The person who wrote the fact-checking article.
    
    \item \emph{author\_website}: This attribute is a link to the fact-checker's profile on the corresponding  news website. The website generally list down articles fact-checked by that individual.
    
    \item  \emph{author\_twitter\_handle}: The Twitter username of the fact-checker is mentioned in this attribute.  
    
    \item \emph{author\_desc or author\_info}:  This attribute gives you a description about the fact-checker.
    
    \end{itemize}

\textbf{Media Features}
The claim viral on any social media platform or mainstream media have many modality. Similarly, the investigation carried out to conclude the status of any viral news is also backed by a numerous number of supporting claims  that can again be in any multimedia form. The set of attributes that are categorized into multimodal features are as follows:

\begin{itemize}
    \item \emph{top\_image}: The first image that is present at the very beginning of the fact-checking article. It generally show the picture present in the viral claim.
    
    \item \emph{image\_links}: The links of all other images that  will either belong to the original claimed images group or are presented in support to the viral claim are put under this feature as a list object. 
    
    \item \emph{video\_links}: For those samples where prime media used for fabrication is video, the link to the original video is provided by fact-checkers to backed their investigation. This attribute stores all such links.
    
    \item \emph{audio\_links}: All the supporting audio links related to the viral claim are presented in this attribute.
    
    \item \emph{links\_in\_text}: To provide complete justification to what was said in the investigation section of the report, authors provide different media links in support of their investigation. All such links are piled in this attribute.
    But to identify, where a specific link is mentioned in the fact-checked article, an attribute named as, \emph{bold\_text} is used for easy identification and matching of the corresponding text from the article.
\end{itemize}

\textbf{Social Features}
The attribute stores the tweet ids present in the sample. The tweet ids can be the post that, \emph{(i)} needs to be investigated or, \emph{(ii)} is present in the support of the fake claim. With this attribute, we can extract the complete information from the tweet thread.

\textbf{Event Features} The set of features in this group gives information about the event to which a news sample belong to. These include \emph{domain} and \emph{tags} attributes. \emph{For example}, the Boom article titled: `False: Chinese Intelligence Officer Reveals Coronavirus Is A Bioweapon' had the following tags (\emph{Coronavirus China, COVID-19, Coronavirus outbreak, Bioweapon, Biological warfare, China, Intelligence Officer}) associated with it. This kind of information is helpful in identifying the genre of the article.

\section{Data Annotation}
In this section, we address the three key questions that facilitated the data annotation process.

\subsection{Need to perform Annotation}
The need to do annotations were two fold. First, we want to highlight the key pieces of extracted text to underline its usability. Since the extracted text neither showed a particular pattern for automatic extraction nor had any distinction between the different textual features discussed in Section \ref{attributes}, we performed manual annotations to excavate the key pieces from the data dump. Second, there was a need to scrutinize each sample to make sure that the proposed dataset do not contain samples, \emph{(i)} that were investigated to be true, \emph{(ii)} articles containing general fact information that news websites usually publish\footnote{https://www.boomlive.in/technologies-will-tackle-irrigation-inefficiencies-agricultures-drier-future/} and, \emph{(iii)} weekly-wrap up articles that increase the chance of deduplication in the dataset.

\subsection{Annotation Process} \label{annotation}

We chose against crowd-sourcing the annotation process 
due to  non-trivial nature of the sub-tasks. We hired individuals who were well versed with the any of the low-resource languages mentioned earlier. They were in the age group of \emph{19-45}. Due to unavailability of language professionals, we performed the annotation process for regional languages with a single entity, whereas two professionals were hired each for lingua franca (\emph{i.e.} Hindi and English). The annotators were first provided with the annotation guidelines which included instructions about each sub-task, definition of the attributes that need to derived from the text and few examples. They studied this document and worked on a few examples to familiarize themselves with the task. They were given feedback on the sample annotations, which helped them to refine their performance on the remaining subset. The annotation process was broken down into two sub-tasks.  In the first sub-task, annotators have to segregate the text into two attributes namely, \emph{claim} and \emph{investigation}. In the second sub-task, annotators have to make sure to chose samples that were sentenced to be false by fact-checkers.

\begin{table}[]
\caption{Inter-annotator agreement for the two sub-tasks. The values in bold indicates that Gwet`s AC(1) and AC(2) scores were calculated for the samples. }
\label{agreement_score}
\begin{tabular}{|c|c|c|c|c|}
\hline
\textbf{Website}                & \textbf{Language} & \multicolumn{2}{c|}{\textbf{Inter Annotator Agreement Score}}                                                                                                          & \textbf{\# Samples} \\ \hline
                                &                   & \begin{tabular}[c]{@{}c@{}}Sub Task 1\\ (Percent  Agreement)\end{tabular} & \begin{tabular}[c]{@{}c@{}}Sub Task 2\\ (Cohen`s Kappa /Gwet`s \\ AC(1) AC(2)\end{tabular} &                     \\ \hline
{Alt News}       & English           & 0.78                                                                      & 0.48                                                                                       & 2058                \\ \cline{2-5} 
                                & Hindi             & 0.76                                                                      & 0.53                                                                                       & 1758                \\ \hline
{Boom}           & English           & 0.42                                                                      & 0.66                                                                                       & 909                 \\ \cline{2-5} 
                                & Hindi             & 0.90                                                                      & 0.53                                                                                       & 880                 \\ \hline
DigitEye                        & English           & 0.86                                                                      & 0.56                                                                                       & 147                 \\ \hline
FactChecker                     & English           & 0.31                                                                      & 0.15                                                                                       & 156                 \\ \hline
{Fact Crescendo} & English           & 1.00                                                                      & \textbf{1.00}                                                                              & 256                 \\ \cline{2-5} 
                                & Hindi             & 0.99                                                                      & \textbf{1.00}                                                                              & 264                 \\ \hline
Factly                          & English           & 0.92                                                                      & 0.76                                                                                       & 971                 \\ \hline
India Today                     & English           & 0.95                                                                      & 0.44                                                                                       & 788                 \\ \hline
News Mobile                     & English           & 0.71                                                                      & 0.29                                                                                       & 1543                \\ \hline
{Vishvas News}   & English           & 0.94                                                                      & \textbf{0.91}                                                                              & 254                 \\ \cline{2-5} 
                                & Hindi             & 0.98                                                                      & \textbf{0.90}                                                                              & 1369                \\ \hline
{Webqoof}        & English           & 0.86                                                                      & 0.47                                                                                       & 1771                \\ \cline{2-5} 
                                & Hindi             & 0.95                                                                      & \textbf{0.97}                                                                              & 328                 \\ \hline
\end{tabular}
\end{table}

\subsection{Annotation Evaluation Metric}
There were two sub-task given to the annotators. To evaluate the performance on the first task, a  customized metric was prepared. For second task, inter-annotator agreement score using Cohen's Kappa and Gwet's AC(1) and AC(2) statistic \cite{gwet} was considered. Next, we will discuss in detail these evaluation metric.

For the first sub-task, annotators were required to do a close reading of the text to extract meaningful information and place it under correct header (\emph{i.e.} claim or investigation). To evaluate the performance, we checked for matched ordinal positions in each annotated piece. A counter value is kept to calculate the number of mismatches. Having tested multiple possible values, we chose a threshold value, d=6 and if d goes beyond the chosen value, then that sample is sent in the conflict resolution bin. The final inter agreement score  is computed using the percent agreement for two raters \cite{percentage}. It is calculated by dividing total count of matched sample with the total number of samples in the data.

For the second sub-task, we evaluated inter-annotator agreements using Cohen's Kappa \cite{cohen}. We observe a mix of moderate and substantial agreement for most of the task. Table \ref{agreement_score} summarizes the Cohen's kappa measures for the sub-task. Though Cohen's Kappa performs exceptionally well when dichotomous decision is involved and takes care of chance agreement too but it fails badly when annotators show near to 100\% agreement. This phenomenon is termed as `the paradoxes of Kappa'. During our evaluation, we observe high agreement between annotators for \emph{1000} samples. To solve the `the paradoxes of Kappa' issue, we used Gwet's AC(1) and AC(2) statistic \cite{gwet}. It overcomes the paradox of high agreement and low reliability coefficients. Table \ref{agreement_score} summarizes the Gwet's score for these samples.

\textbf{Conflict Resolution} We chose a straightforward approach for adjudication in case of disagreements in all the sub-tasks. If the two annotators present a contradictory views for a  sample, the annotations for all the disagreed tasks are then adjudicated by verbal discussions among them.

\section{Basic Dataset Characterization}
We begin by providing  a statistical overview of our proposed dataset.

\subsection{Summary Statistics}

\begin{figure}[h]
  \centering
  \includegraphics[width=\linewidth]{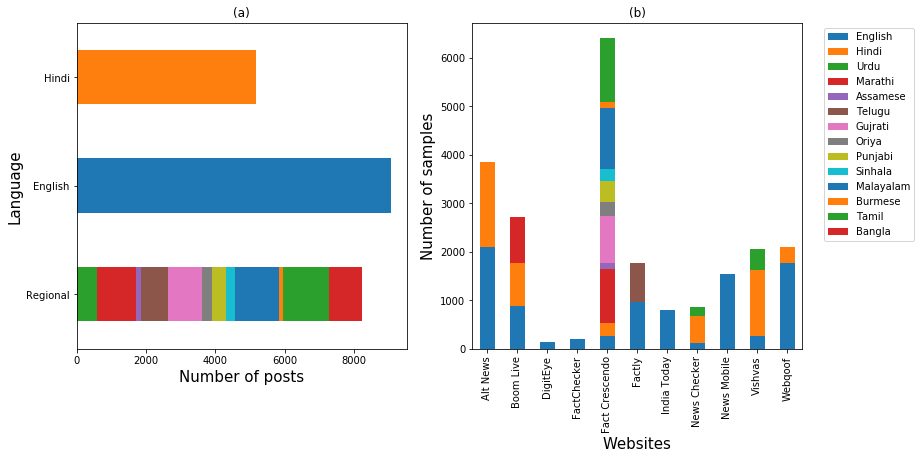}
  \caption{\emph{(a)} shows the distribution of different languages in our proposed dataset, \emph{(b)} shows the spread of data across different websites. It also depicts the languages supported by each website.} 
  \label{langauge}
\end{figure}

Figure \ref{langauge}\emph{(a)} shows the distribution of samples across languages in our proposed \emph{FactDRIL}. Surprisingly, \emph{Regional} languages surpasses \emph{Hindi} by \emph{37\%}. The diffusion of samples in the regional interface is majorly dominant by \emph{Bangla}, \emph{Malayalam}, \emph{Urdu} and \emph{Marathi} langauge. Figure \ref{langauge}\emph{(b)} represents the number of samples belonging to the \emph{eleven} fact-checking websites. Among them \emph{Fact Crescendo} website rules in debunking fake news dissemination in different languages.

\subsection{Popular Fake Events in India}
\begin{figure*}[h]
\includegraphics[width=4.6cm]{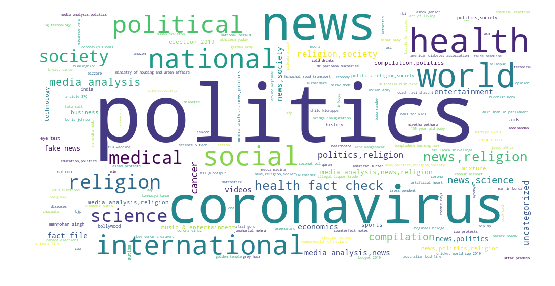}
\includegraphics[width=4.5cm]{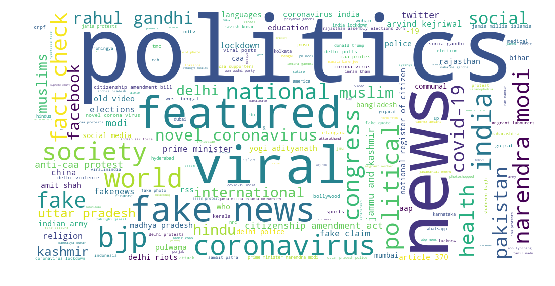}
\includegraphics[width=4.6cm]{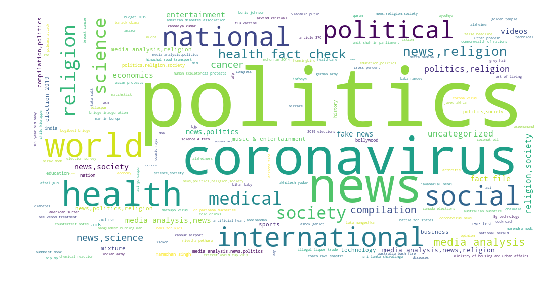}
\caption{Topic Distribution in English, Hindi and Regional languages (left to right). All the figures clearly shows that the majority of the fake news dissemination across the country is centered towards political domain.}
\label{domain}
\end{figure*}

We analyze the topic distribution of fact-checking articles in different languages, \emph{i.e.} English, Hindi and Regional languages. From Figure \ref{domain} (a), (b), (c), we can conclude that political activity is an important ground for fake news creation. With the onset of \emph{2020}, the world has witnessed a global pandemic \emph{i.e.} Coronavirus. This has not only affected lives of people but has also given rise to infodemic of misinformation. To no surprise, the second popular domain for fake news creation in India was Coronavirus followed by health and religion. 

\subsection{Circulation of Fake News in India}
\begin{figure}[h]
  \centering
  \includegraphics[width=\linewidth]{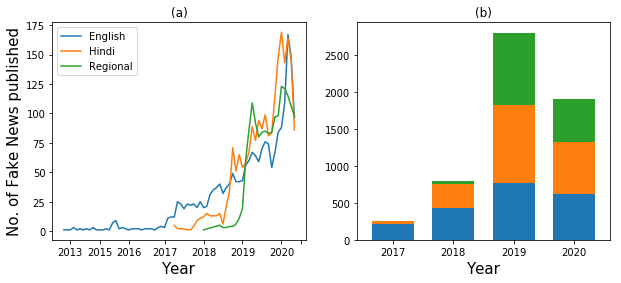}
  \caption{Circulation of fake news over the years in India. For the year \emph{2020}, the data is collected till \emph{June 2020}.}
  \label{year_wise}
\end{figure}

Figure \ref{year_wise}\emph{(a)} shows that the fact-checking trend came to India in 2012, majorly debunking news in English Language.  As and when fake news dissemination in English got little popular \emph{i.e.} 2017, we saw it intruding in the other languages too. This steady shift to other languages was observed quite lately \emph{i.e.} in 2017 and 2018. We observe sharp peaks and drops in the graph that will be an interseting study to do in future. Figure \ref{year_wise}\emph{(b)} shows the year-wise distribution of samples in the dataset. The graph shows a steady increase in fake news creation over the years with a major peak observed in 2019. 
For both these observations, the data considered for the year \emph{2020} is till June.

\section{Use Cases}

There are varied threads of misinformation research
that can be initiated with the help of FactDRIL. We would like to formally propose some ideas for the same.
\begin{itemize}
    \item Improve Misinformation Detection System: Till date, various efforts have been made to eliminate misinformation from the ecosystem. The major drawback observed in such methods is two fold. First, the system performs well on trained samples and fails drastically for the real-world data. Secondly, the performance of classifiers varies considerably based on the evaluation archetype and performance metric \cite{Bozarth_Budak_2020}. We present a dataset that provides a detailed investigation of the fake sample that includes, \emph{(i)} the modality faked in the news, \emph{(ii)} `how' the sample was concluded to be false and, \emph{(iii)} tools used to draw that conclusion. We believe that a detailed study in understanding the formation of fake news pattern can improve the performance of classifiers to tackle misinformation in the real world.
    
    \item Suppressing Fake News Dissemination at an Early Stage: We all know that fake news is not new, probably it is as old as humanity. Despite constant efforts made to eliminate it from the ecosystem, it finds a way to dissolve in our lives. The impact of fake information is devastating. We believe eradicating fake news is bit challenging but creating systems that can suppress its effect is a feasible task. With \emph{FactDRIL}, we can develop technologies that can be stationed at different social media platforms.  Such system can use information from the debunked pieces and stop proliferation of its variants on the platform.
    
    \item Bias among fact-checkers: Fact-checking is tedious. Different websites aim to debunk news of different genre. There can be websites that aims at exposing a particular kind of information. It will be interesting to look out for biases in the fact-checking pattern and its related effects.
    
    \end{itemize}

\section{Conclusion}
In this paper, we presented \textbf{FactDRIL}: a \textbf{Fact}-checking \textbf{D}ataset for \textbf{R}egional \textbf{I}ndian \textbf{L}anguages. To our knowledge, this is first large scale multi-lingual Indian fact-checking data that provide fact-checked claims for low resource languages. The dataset can be accessed on the request via \href{https://bit.ly/3btmcgN}{\underline{link}}.\footnote{https://bit.ly/3btmcgN} We believe such datasets will allow researchers to explore the fake news spread in regional languages. Additionally, researchers could also look out for dissemination of fake content across the different language silos. The dataset comprises of \emph{22,435} samples crawled from the \emph{11} Indian fact-checking websites with IFCN certification. A vast range of low-resource languages has been covered including, Bangla, Marathi, Malayalam, Telugu, Tamil, Oriya, Assamese, Punjabi, Urdu, Sinhala and Burmese. The features curated from the data dump is further grouped under \emph{meta}, \emph{textual}, \emph{media}, \emph{social}, \emph{event} and \emph{author} features.  We also present a new attribute to the feature list \emph{i.e. investigation reasoning} and explain its relevance and need in the current fact-checking mechanism. Currently, this feature is extracted via manual intervention. In future, we plan to automate the attribute extraction from the text. We would also like to organize challenges around this data to instigate researchers in asking interesting questions, find limitations and propose any improvements or novel computational techniques in detecting fake news in low-resource languages.








\bibliographystyle{ACM-Reference-Format}
\bibliography{sample-base}

\appendix









\end{document}